\documentclass[conference]{IEEEtran}
\IEEEoverridecommandlockouts
\usepackage{cite}
\usepackage{amsmath,amssymb,amsfonts}
\usepackage{algorithmic}
\usepackage{graphicx}
\usepackage{textcomp}
\usepackage{xcolor}
\DeclareUnicodeCharacter{2061}{}
\def\BibTeX{{\rm B\kern-.05em{\sc i\kern-.025em b}\kern-.08em
    T\kern-.1667em\lower.7ex\hbox{E}\kern-.125emX}}
\begin{document}

\title{Learning Invariant Representations for Equivariant Neural Networks Using Orthogonal Moments}
\author{\IEEEauthorblockN{1\textsuperscript{st} Jaspreet Singh}
\IEEEauthorblockA{\textit{Department of Computer Science} \\
\textit{Punjabi University}\\
Patiala, India \\
jaspreet{\_}rs21@pbi.ac.in}
\and
\IEEEauthorblockN{2\textsuperscript{nd} Chandan Singh}
\IEEEauthorblockA{\textit{Department of Computer Science} \\
\textit{Punjabi University}\\
Patiala, India \\
chandan.csp@gmail.com}
}

\maketitle

\begin{abstract}
The convolutional layers of standard convolutional neural networks (CNNs) are equivariant to translation. However, the convolution and fully-connected layers are not equivariant or invariant to other affine geometric transformations. Recently, a new class of CNNs is proposed in which the conventional layers of CNNs are replaced with equivariant convolution, pooling, and batch-normalization layers. The final classification layer in equivariant neural networks is invariant to different affine geometric transformations such as  rotation, reflection and translation, and the scalar value is obtained by either eliminating the spatial dimensions of filter responses using convolution and down-sampling throughout the network or average is taken over the filter responses. In this work, we propose to integrate the orthogonal moments which gives the high-order statistics of the function as an effective means for encoding global invariance with respect to rotation, reflection and translation in fully-connected layers. As a result, the intermediate layers of the network become equivariant while the classification layer becomes invariant. The most widely used Zernike, pseudo-Zernike and orthogonal Fourier-Mellin moments are considered for this purpose. The effectiveness of the proposed work is evaluated by integrating the invariant transition and fully-connected layer in the architecture of group-equivariant CNNs (G-CNNs) on rotated MNIST and CIFAR10 datasets.
\end{abstract}


\section{Introduction}
\label{sec:intro}
Recently, deep convolutional neural networks (CNNs) have achieved new state-of-the-art accuracy for various computer vision applications including visual object recognition. Among the other factors, convolutional weight sharing and depth are the two most important factors behind the remarkable success of CNNs \cite{b13}. Generally, the architectures of CNNs are consist of two parts: the feature extraction part and classification part. The convolutional layer is the core building block of the feature extraction part that does most of the computation and responsible to learn the abstractions present in the input data. The convolutional layers of CNNs are equivariant to translation which means shifting the original image and then feeding through the network is similar to first feeding the original input image and then shifting the feature maps \cite{b23}. However, the standard CNNs are not equivariant to other affine geometric transformations such as rotation and reflection. Invariance or equivariance with respect to different geometric transformations is one of the highly desirable properties of the deep learning models, especially for the task of image classification. As a result, a novel class of CNNs is proposed which are equivariant to different affine geometric transformations (i.e., rotation, reflection, and translation) either by utilizing the group-equivariant convolutional operators  \cite{b3} or steerable filters. Generally, the conventional layers and standard operations (i.e., convolution, pooling, batch normalization and activation functions) in CNNs are replaced with equivariant layers and operations in equivariant neural networks. The fully-connected layer is neither equivariant nor invariant to any transformations, consequently, not used in equivariant networks. Almost all the equivariant networks are either conditioned to perform convolution and down-sampling over the filter responses until the spatial dimensions get eliminated or average is taken over the filter responses to get the scalar fields for classification layer. 

\textbf{Contribution} we propose a simple but very effective solution by integrating the circular continuous orthogonal moments into a transition between the convolutional and fully-connected layers in equivariant neural networks to encode the global invariance with respect to rotation, reflection and translation instead of down-sampling filter responses to eliminate the spatial dimensions or taking the average to encode invariance in the final layer\cite{b3}. As a result, the intermediate layers of the network become equivariant to different transformations while fully-connected layer becomes invariant. The Zernike, pseudo-Zernike and orthogonal Fourier-Mellin moments are considered to encode invariance which provides higher-order statistics of the input function instead of using average or maximum of input function.  Discrete Fourier transforms (DFTs) has been successfully used in the design of equivariant convolutional layers \cite{b9}, as of yet, the applications of orthogonal moments to equivariant networks has been relatively overlooked  \cite{b6, b17}. In contrast to DFTs, the kernel functions of orthogonal moments are real polynomial functions and one can compute infinite number of moments.  
\section{Related Work}
\label{sec:rwork}
The conventional hand-crafted feature descriptors used in the area of computer vision are broadly classified into local and global descriptors. Moments are one of the most popular feature descriptors which belong to the class of global shape descriptors \cite{b25}. The popularity of moments among the various global descriptors is due to their invariance properties with respect to different affine geometric transformations and their robustness towards  noise \cite{b17}. Thus, moments have been successfully applied to various computer vision applications such as image matching, denoising, image classification, segmentation, etc. The class of continuous orthogonal moments which are defined in the polar domain are used more commonly as compared to their non-orthogonal counterparts and these moments are popularly known as orthogonal rotation invariant moments (ORIMs) \cite{b12}. Among the number of moments that belong to the class of ORIMs, the Zernike moments (ZMs), pseudo-Zernike moments (PZMs), and orthogonal Fourier-Mellin moments (OFMMs) are used more regularly. 

It is important to distinguish between the terms invariance and equivariance because these terms are used frequently throughout this paper. A feature descriptor is invariant to a given transformation (i.e., rotation, reflection and translation) when the output feature vector does not change given the transformed input. On the other side, a feature descriptor is equivariant with respect to a given transformation when the output transforms in a predictable manner with respect to transformed input. The convolutional layers of the standard CNNs are only equivariant to translation. Cohen and Welling\cite{b3} in their seminal work proposed a new class of convolutional neural networks known as group-equivariant CNNs (G-CNNs) based on the group theory. The convolution layer of the standard CNN is replaced with group convolution layer in G-CNN. The convolution layer in CNN is a special case of group convolution layer and group convolution layers are the only layers in the linear neural networks that are guaranteed to be equivariant. The key feature of G-CNNs is its equivariance with respect to the transformations defined by the special group. However, G-CNNs are equivariant to discrete transformations that leave the pixel grid intact (e.g., 90°-rotations, translations and reflections). Hoogeboom et al.\cite{b7} proposed the HexaConvs which has 6-fold rotational symmetry as compared to the 4-fold rotational symmetry of G-CNNs which allows more parameter sharing. The proposed HexaConvs are able to utilize symmetry equivariance and invariance which is vital factor behind its better performance as compared to other techniques. Bekkers\cite{b1} proposed a modular framework for G-CNNs for arbitrary Lie groups in order to overcome the limitations of standard G-CNN \cite{b3} which is practically applicable to only either discrete groups or continuous compact groups. Furthermore, Lafarge et al.\cite{b10} proposed the $SE(2)$ group convolutional operator which proves that a concatenation of two roto-translations results in a net roto-translation. The three new layers are introduced to achieve fully equivariance throughout the CNN: a lifting layer, group-convolution layer and a projection layer. As a result the proposed $SE(2)$ equivariant G-CNN is not only equivariant to orientations in the input data that lay on the pixel gird but also to orientations that are out of the pixel grid. The key features of $SE(2)$ G-CNNs are: i) it learns the geometric structures into the network architecture and ii) equivariance is guaranteed. Chidester et al.\cite{b2} proposed a new equivariant convolutional scheme known as conic convolution which is an alternative to group-convolution. In the case of conic convolution, the rotated filters are convolved only over the conic regions of the input feature maps rather than across the entire image. The proposed technique is computationally efficient and provides better performance. Worrall et al.\cite{b23} proposed the Harmonic networks (H-Nets) by restricting the filters of the convolutional layers to be from the circular harmonic family. H-Net hard-bake patch-wise 360° rotational equivariance into deep image representation. The theory of steerability reveals that a steerable filter can be constructed at any rotation as a linear finite combination of base filters which eliminates the need to learn multiple filters for different rotation angles. The new class of efficient and flexible equivariant CNNs are developed based on the steerable filters which has achieved state-of-the-art performance on the standard image classification datasest \cite{b4,b21}. Weiler et al.\cite{b22} developed a Steerable Filter CNNs (SFCNNs) which is equivariant under translation and rotation. The key property of the SFCNNs is that it learns the steerable filters by avoiding the interpolation artifacts. Further, SFCNNs utilize group convolution in the intermediate layers to ensure an equivariant mapping of feature maps. In addition to using group theory or steerable filters to design the equivariant layers, Sabour et al.\cite{b15} proposed the capsule neural networks. Each capsule in the capsule network is a group of convolutional neurons and the dynamic routing algorithm is developed for learning between the primary and digit capsules. Capsule networks are equivariant to complex global transformations.
\section{CNNs and Group-Equivariant CNNs}
\label{sec:CNNs}
Let $l$ be a particular layer in a CNN model, the feature map $f$ and filter $\phi$ of dimension $K$ are represented by $f:\mathbf{Z}^2\rightarrow \mathbf{R}^K$ and $\phi:\mathbf{Z}^2\rightarrow \mathbf{R}^K$, respectively. The convolution operation $(\ast)$ is defined as follows:
\begin{equation}
\label{eq:1}
    f\ast \phi(x) = \sum_{z\in \mathbf{Z}^2}\sum_{k=1}^{K^l}f_k(z)\phi_k(x-z).
\end{equation}
The standard convolutional operation in CNN is equivariant to translation. However, it is not equivariant to other affine geometric transformations such as rotation and reflection \cite{b3}. Let $L_r$ be a operator which rotates a feature map $f$ by $r$ then the convolution of a rotated $f$ with $\phi$ is equals to the rotation of convolution between $f$ and inversely rotated filter $L_{r^{-1}}\phi$ given as follows \cite{b7}:
\begin{equation}
    \label{eq:r1}
    [[L_{r}f]\ast\phi](x)=L_r[f\ast[L_{r^{-1}}\phi]](x).
\end{equation}
Since $[L_{r}f]\ast \phi$ can not be expressed in terms of $f\ast \phi$, thus, convolution is not rotation equivariant.

Cohen and Welling\cite{b3} generalized the convolutional operation to operate on functions on groups in order to achieve equivariance with respect to other transformations. Mathematically, a group is a set combined with a binary operation which together follows the conditions of identity, inverse, associativity and closure. Let $G$ be a group and $f\in \mathbf{Z}^2$ is an input image then the first convolutional layer in group-CNN (G-CNN) is defined as follows \cite{b3}:
\begin{equation}
\label{eq:2}
    f\ast \phi(g) = \sum_{z\in Z^2}\sum_{k=1}^{K}f_k(z)\phi_k(g^{-1}z),
\end{equation}
where $g\in G$ is a transformation about the origin (e.g. rotation and reflection). In the case of standard convolutional operation given in \eqref{eq:1}, the filter is translated over the image and the inner product is computed at each translation, while in group-convolution, the filter is transformed by each element of $G$. The output of group convolution operation defined in \eqref{eq:2} is a function on the group $G$. The group convolution operation in subsequent layers of G-CNN must operate on group functions which are defined as follows:
\begin{equation}
    f\ast \phi(g) = \sum_{h\in G}\sum_{k=1}^{K}f_k(h)\phi_k(g^{-1}h).
\end{equation}
The standard operation of neural networks including pooling, batch normalization, and activation functions are redefined for group functions to preserve the equivariance property \cite{b3}.
\section{Mathematical Framework of Continuous Orthogonal Moments}
\label{sec4}
Let $O$ be a function defined in the continuous polar domain $(r,\theta)$. Then continuous orthogonal moments of order $p$ and repetition $q$ over the unit disk for the function $O$ are defined as follows \cite{b8}:
\begin{equation}
\label{eq:4}
    OM_{p,q}(O)=\lambda_ p\int_{0}^{2\pi}\int_{0}^{1}O(r,\theta)R_{p,q}(r)e^{-iq\theta}rdrd\theta,
\end{equation}
where $i=\sqrt{-1},$ $p\in Z^+,$ $q\in Z,$  $\lambda_p$ is the normalization parameter, and $R_{p,q}(r)$ is the radial polynomial basis function. The moments which belong to the class of orthogonal moments and defined in the continuous polar domain differ only in the form of their radial basis function $(R_{p,q} (r))$. The radial basis functions $(R_{p,q} (r))$ of the three most commonly used moments which are Zernike moments (ZMs) \cite{b20}, pseudo-Zernike moments (PZMs) \cite{b24} and orthogonal Fourier-Mellin moments (OFMMs) \cite{b18} are shown in Table \ref{ta:1}. The orthogonal moments which are defined in the continuous polar domain are also known as orthogonal rotation invariant moments (ORIMs) because the magnitude of these moments is invariant to rotation and reflection \cite{b19}.
\begin{table*}
\caption{Radial basis functions $(R_{p,q} (r))$ along with normalization parameter $\lambda_p$ of ZMs, PZMs, and OFMMs with the conditions on the parameters $p$ and $q$, and positive number of moment coefficients for $p_{max}$.}
\begin{center}
\begin{tabular}{|l|c|c|c|}
\hline
\bfseries Moments & \bfseries $\lambda_p$ & \bfseries Radial function
$R_{p,q}(r)$ & \bfseries Number of moment \\
 & & & \bfseries coefficients for $p_{max}$\\
\hline\hline
ZMs \cite{b20} &$\frac{p+1}{\pi}$& $\sum_{k=0}^{\frac{p-|q|}{2}}\frac{(-1)^k(p-k)!}{k!\left(\frac{p+|q|}{2}-k\right )!\left(\frac{p-|q|}{2}-k\right )!}r^{p-2k}$ & $\frac{1}{8}\Big[2p_{max}^2+4p_{max}+$\\
 & & $|q|\leq p, p-|q|=even$ &$(-1)^{2p_{max}}+(-1)^{p_{max}+1}\Big]$\\
\hline
PZMs \cite{b24} &$\frac{p+1}{\pi}$& $\sum_{k=0}^{p-|q|}\frac{(-1)^k(2p+1-k)!}{k!\left(p-|q|-k\right )!\left(p+|q|+1-k\right )!}r^{p-k}$ &$\frac{1}{2}p_{max}(p_{max}+1)$\\
 & & $|q|\leq p$&\\
\hline
OFMMs \cite{b18} &$\frac{p+1}{\pi}$& $\sum_{k=0}^{p}\frac{(-1)^{p+k}(p+k+1)!}{k!\left(k+1\right )!\left(p-k\right )!}r^{k}$ &$\frac{1}{2}p_{max}(p_{max}+1)$\\
& & $|q|\leq p$&\\
\hline
\end{tabular}
\end{center}
\label{ta:1}
\end{table*}
The moments defined using \eqref{eq:4} is for the continuous functions in the polar coordinate system $(r,\theta)$ over the unit disk. However, digital computers work with the discrete functions defined in the cartesian domain. Let $O(s,t)$ be a discrete function defined in the cartesian domain of size $M\times M$, where $(s,t)\in[0,M-1]\times[0,M-1]$, then a mapping from cartesian to polar domain is performed. Let $(s,t)$ be a location in the cartesian domain, then its corresponding coordinates in the polar domain $(r_{st},\theta_{st})$ are derived using $r_{st}=\sqrt{x_s^2+y_t^2 }$ ($x_s$ and $y_t$ are defined below), and $\theta_{st}=\tan^{-1}⁡(y_t/x_s)$, where $\theta_{st}\in[0,2\pi]$. The condition $x_s^2+y_t^2\leq1$ is imposed to restrict the computation on the unit disk.

The following transformation is used to map the coordinates of $(s,t)$ of a discrete function of size $M\times M$ into a unit disk \cite{b20}:
\begin{equation}
\begin{split}
\label{eq:5}
    x_s=\frac{2s+1-M}{D},\ y_t=\frac{2t+1-M}{D}, \\   s,t=0,1,2,\dots M-1,
\end{split}
\end{equation}
and

$$
D= \left\{
  \begin{array}{lr} 
      M & for\ inner\ unit\ disk, \\
      M\sqrt{2} & for\ outer\ unit\ disk, 
      \end{array}\right\}
$$
where $x_s$ and $y_t$ represent the cartesian coordinates $(s,t)$ on the unit disk in the polar domain. The elemental area occupied by each coordinate is $\Big[x_s-\frac{\Delta x}{2}, x_s+\frac{\Delta x}{2}\Big] \times \Big[y_t-\frac{\Delta y}{2},\ y_t+\frac{\Delta y}{2}\Big]$, where $\Delta x= \Delta y=\frac{2}{D}$. It may be observed that a digital coordinate $(s,t)$ is mapped to a location $(x_s,y_t)$ on the unit disk by translating the origin $(0,0)$ to the center $\left(M/2,M/2\right)$, and then scaling the resulting values by the scaling factor $\lambda=\frac{2}{D}$. This mapping ensures that resulting coordinates $(x_s,y_t )\in[-1,1]\times[-1,1]$ and the condition $x_s^2+y_t^2\leq1$ ensure that the computations are performed inside the unit disk. 

Since there is no direct analytical solution to the double integration given in \eqref{eq:4} for the radial basis functions of ORIMs given in Table \ref{ta:1}. The zeroth-order approximation of \eqref{eq:4} is commonly used which is defined as:
\begin{equation}
\label{eq:6}
    OM_{p,q}(O)=\lambda_p\sum_{s=0}^{M-1}\sum_{t=0}^{M-1}O(x_s,y_t)R_{p,q}(r_{s,t})e^{-iq\theta_{st}}.
\end{equation}
It is important to mention here that $OM_{p,q}(O)$ is complex, and ORIMs are implemented efficiently using vectorization and the computational complexity of computing each moment coefficient $(p,q)$ for ZMs, PZMs and OFMMs is $O(p_{max})$.
\section{Invariance Properties of Moments}
\label{sec5}
In the following subsections, we discuss the invariance properties of ORIMs on rotation, reflection and translation.
\subsection{Rotation Invariance}
\label{RRI}
Let $O(r,\theta)$ be a function and an arbitrary angle $\alpha$, where $\alpha \in[0^\circ,360^\circ)$. If the function $O(r,\theta)$ is rotated by an angle $\alpha$ in the counter clockwise direction around its center, then a pixel at location $(r,\theta)$ is shifted to $(r,\theta+\alpha)$. Therefore, $O^\alpha (r,\theta)=O(r,\theta+\alpha)$. The moments of the rotated function $OM_{p,q} (O^\alpha )$ and the unrotated function $OM_{p,q} (O)$ has the following relationship (see Appendix A for details)\cite{b14}:

\begin{equation}
\label{eq:7}
    OM_{p,q}(O)=OM_{p,q}(O^\alpha)e^{-iq\alpha}.
\end{equation}
 This relationship shows that the moments of the original and rotated image undergo phase-shift by an angle $- q\alpha$ and the magnitude of the moments remain the same.  It is quite straightforward to achieve the invariance with respect to rotation from this relationship by taking the magnitude on both sides which cancels the role of phase angles as follows:
\begin{equation}
\label{eq:8}
    \left|OM_{p,q}(O)\right|=\left|OM_{p,q}(O^\alpha)\right|.
\end{equation}
The magnitude of ORIMs is invariant to any arbitrary rotation angle $\alpha$ in $[0^\circ, 360^\circ)$.
\subsection{Reflection Invariance}
\label{RI}
Let $O^{hf} (s,t )=O(-s,t)$ is the horizontal and $O^{vf} (s,t )=O(s,-t)$ is the vertical flipped versions of the discrete function $O(s,t)$. The relationship between the moments of the flipped function $OM_{p,q} (O^{hf})$ and the original function $OM_{p,q} (O)$ is defined as follows (see Appendix B for details)\cite{b12}:
\begin{equation}
\label{eq:9}
    OM_{p,q}(O^{hf})=(-1)^q OM_{p,q}^{\ast}(O).
\end{equation}
Similary, the relationship between the moments of the vertically flipped function $OM_{p,q} (O^{vf})$ and the original function $OM_{p,q} (O)$ is defined as follows:
\begin{equation}
\label{eq:10}
    OM_{p,q}(O^{vf})=OM_{p,q}^{\ast}(O),
\end{equation}
where $OM_{p,q}^{\ast}(O)$ is the complex conjugate of $OM_{p,q}(O)$. Again the magnitudes of \eqref{eq:9} and \eqref{eq:10} are invariant to both horizontal and vertical flipping.
\subsection{Translation Invariance}
\label{TI}
In Section \ref{sec4}, the moments are computed using \eqref{eq:6} by mapping the center $(M/2, N/2)$ of function $O$ to the origin of unit disk. Invariance to translation can be achieved by mapping the function $O$ such that the centroid of $O$ coincide with the origin of unit disk. The central moments invariant to translation are computed as follows (see Appendix C for details)\cite{b5}:
\begin{equation}
\label{eq:11}
    \overline{OM}_{p,q}(O)=\lambda_p\sum_{s=0}^{M-1}\sum_{t=0}^{M-1}O(\overline{x}_s,\overline{y}_t)R_{p,q}(\overline{r}_{s,t})e^{-iq\overline{\theta}_{st}}.
\end{equation}
It is important to note here that moment function $OM_{p,q}(O)$ is defined in the complex space and magnitude of central moments $\overline{OM}_{p,q}(O)$ is taken which is a real value to make the moments invariant to rotation and reflection.
\section{Invariant Transition using ORIMs}
\label{sec6}
In standard CNNs, some number of fully-connected layers are applied after the final convolution layer to combine the filter responses. However, the layers and standard operations (i.e., convolution, pooling, batch normalization and activation functions) in CNNs are replaced with equivariant layers and operations in equivariant networks. Since, fully-connected layer is neither equivariant nor invariant. Thus, either down-sampling over filter responses is performed to eliminate the spatial dimensions or average is taken over the filter responses to get scalar fields for classification layer \cite{b3,b10,b22,b21}. \\
The general framwork of 2-D ORIMs is discussed in Section \ref{sec4}. Since the basis functions of moments are orthogonal, the two basis functions $V_{p,q}(r,\theta)$ and $V_{p',q'}(r,\theta)$ belong to the basis function set are highly uncorrelated
\begin{equation}
    \int_0^{2\pi}\int_0^1 V_{p,q}(r,\theta)V_{p',q'}(r,\theta) rdrd\theta=\frac{1}{\lambda_p}\delta_{p,p'}\delta_{q,q'}
\end{equation}
where $\delta_{p,p'}=1$ if $p=p'$ and $0$ otherwise. Due to the orthogonal basis set, the error estimation is easy when the limited number of projections are given and reconstruction is also simple. Moreover, the kernel function $R_{p,q}(r)$ of different orders $(p,q)$ have different number of zero-crossings and shapes which is very useful to represent the discriminative features of the input function $O(r,\theta)$.
\begin{figure*}
\begin{center}
\centerline{\includegraphics[width=16.0cm,height=4.0
cm]{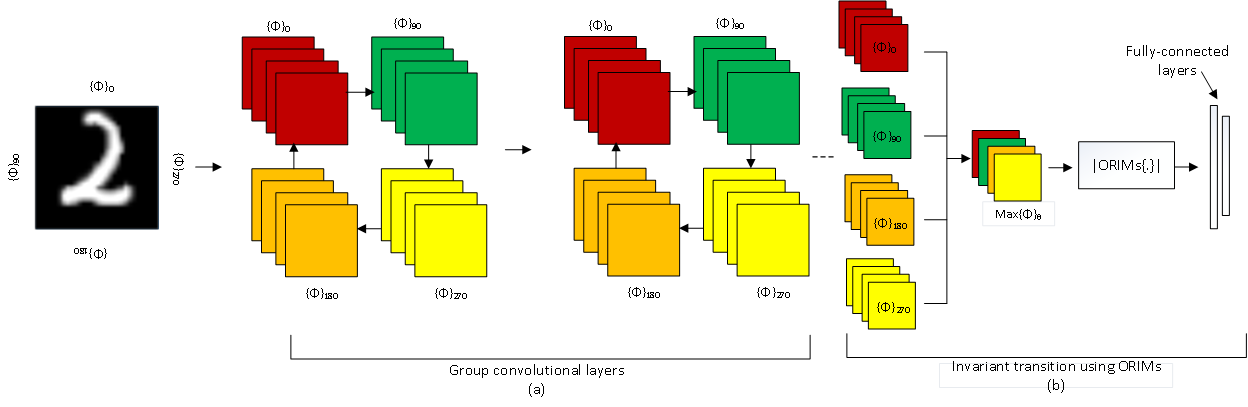}}
\caption{The proposed equivariant and invariant architecture using intermediate group convolution layers and invariant fully-connected layers using ORIMs, respectively.}
\label{fig:1}
\end{center}
\end{figure*}
Instead of eliminating the spatial dimensions of filter responses, the higher-order moment invariant $\overline{OM}_{p,q}(.)$ are computed over the feature maps $(X^l)$ of layer $l$.
\begin{equation}
\label{re:3}
    I_f=\overline{OM}_{p,q}(X^l_f), f=1,2,...,F^l.
\end{equation}
The intermediate layers $l$, where $l>1$, are equivariant to the transformations defined by group G. The feature maps $X^l$ for $l>1$ are function on $G$ in G-CNN and dimensions of $X^l$ are $H^l\times W^l\times \theta \times F^l$, where $\theta$ represent the number of transformations (e.g. 1 for CNN and 4 or 8 for G-CNN). Before applying (\ref{re:3}), the maximum projection is taken over $\theta$ denoted as $max_\theta(X^l)$. As a result, the feature maps $X^l$ becomes function on $\mathbf Z^2$ and the higher-order moment invariants are computed over $X^l$. The resulting invariant features $I$, where $I=[I_1,I_2,...,I_{F^l}]$ are passed to fully-connected layer for classification.

The process of encoding invariance using ORIMs is shown in Figure \ref{fig:1}. It is explicit from the discussions in Sections \ref{sec4} and \ref{sec5} that the ORIMs project the input function on to the orthogonal basis set, consequently, the features are highly discriminative, invariant and represent the complete aspects of the input function. 
\section{Experiments}
\label{sec7}
In the following subsections, the proposed equivariant and invariant architecture is evaluated on rotated MNIST and CIFAR10 datasets. Rotated MNIST and CIFAR10  datasets are chosen because results are reported using G-CNN \cite{b3} on these two datasets. Thus, the effectiveness of the proposed invariant transition is also evaluated using these two datasets and by concatenating the invariant transition followed by fully-connected layer to the architectures used for G-CNNs \cite{b3} experiments\footnote{Source code of G-CNN+ORIMs is available at: https://github.com/JaspreetSinghMaan/G-CNN-ORIMs.} 
\subsection{Rotated MNIST}
The rotated MNIST dataset \cite{b11} consists of 62,000 handwritten digit images which are divided into training, validation and testing sets of size 10000, 2000 and 50000, respectively. It is important to mention here that the train and validation sets are un-rotated while the images of test set are rotated randomly in $[0,2\pi)$. 

\begin{figure*}
\begin{center}
\includegraphics[width=5.5cm,height=4cm]{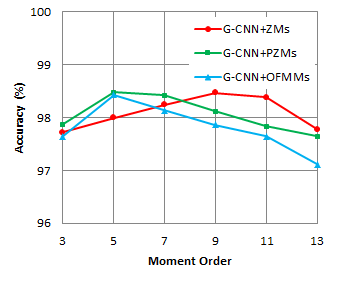}
\includegraphics[width=5.5cm,height=4cm]{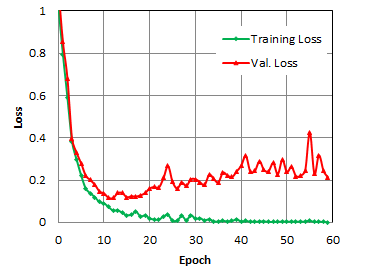}
\includegraphics[width=5.5cm,height=4cm]{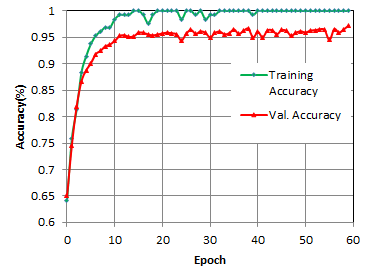}
\caption{(Left) Recognition rates (\%) obtained by G-CNN+ZMs, G-CNN+PZMs, and G-CNN+OFMMs for different moment orders, (Center) training and validation loss obtained by G-CNN+PZMs, and (Right) training and validation accuracy obtained by G-CNN+PZMs.}
\label{fig:2}
\end{center}
\end{figure*}

We trained the network, according to the specifications specified in \cite{b3}. The proposed invariant transition from convolutional to fully-connected is integrated after layer 6 $(l=6)$ of the G-CNN architecture given in \cite{b3}. The output dimensions of the feature maps $(X^l)$ at layer $l=6$ are $4\times4\times8\times10$, where $H^l=4,\ W^l=4,\ \theta=8,$ and $F^l=10$. The maximum projection is taken over the transformation $(\theta)$ axis which results into feature maps $(X^6)$ of dimension $4\times 4\times 10$. It is important to mention here that after integrating the invariant transition followed by a fully-connected layer results into approximately equal number of parameters as in G-CNN. The results obtained by the proposed G-CNN+ORIMs (i.e., G-CNN+ZMs, G-CNN+PZMs, and G-CNN+OFMMs) are shown in Figure \ref{fig:2} in left for the different moment orders $p_{max}$. The purpose of changing the moment orders is to select the optimal moment order $p_{max}$ for different moment functions. The optimal moment order $(p_{max})$ for G-CNN+ZMs is $9$ while for G-CNN+PZMs and G-CNN+OFMMs optimal order is $5$. Further, in this section, only  optimal moment orders $(p_{max})$ are used to perform the experiments. Table \ref{tab2} shows the results obtained by the existing state-of-the-art CNN, CNN+data aug, G-CNN, CFNet, H-Net and the proposed G-CNN+ZMs, G-CNN+PZMs, and G-CNN+OFMMs. The minimum test error is obtained by G-CNN+PZMs of $1.62\%$, followed by G-CNN+ZMs of $1.63\%$ and G-CNN+OFMMs of $1.67\%$ which has reduced the test error (\%) significantly when compared to CNN, CNN+data aug, G-CNN, CFNet and marginally to H-Net. The center and right figures in Figure \ref{fig:2} shows the train-validation loss and accuracy, respectively, obtained by G-CNN+PZMs.

\begin{table}[htbp]
\caption{Test error (\%) obtained on rotated MNIST dataset.}
\begin{center}
\begin{tabular}{c|c|c}
\hline
\textbf{Method}&\multicolumn{2}{|c}{\textbf{Rotated MNIST}} \\
\cline{2-3} 
 & \textbf{\textit{Test error (\%)}}& \textbf{\textit{\#params}}\\
\hline
CNN \cite{b3} & 5.03$\pm$ 0.002 & 22k\\ 
CNN+data aug & 3.50 & 22k \\
G-CNN \cite{b3} & 2.28$\pm$ 0.0004 & 25k\\ 
CFNet \cite{b2} & 2.00 & -\\ 
H-Net \cite{b23} & 1.69 & 33k \\ \hline
G-CNN+ZMs & 1.63$\pm$ 0.003 &25k \\ 
G-CNN+PZMs & 1.62$\pm$ 0.002 &25k \\
G-CNN+OFMMs & 1.67$\pm$ 0.005 &25k \\ \hline
\end{tabular}

\label{tab2}
\end{center}
\end{table}
\subsection{CIFAR10}
CIFAR10 is a color image dataset which consists of 60,000 images categorized into 10 classes and the size of each image is $32\times32$. The dataset is split into training, validation and testing splits of size 40,000, 10,000 and 10,000, respectively. We compare the proposed G-CNN+ZMs, G-CNN+PZMs, and G-CNN+OFMMs with the CNN and G-CNN given in \cite{b3}. Thus, the experiments are conducted according to the model architecture and specifications specified in \cite{b3}. Here, instead of taking the average over the feature maps $(X^l )$ after layer 8 $(l=8)$, the ORIMs-based transition procedure is integrated. The dimensions of the output feature maps at layer 8 $(X^8)$ are $8\times8\times8\times64$ and after taking the maximum projection over the transformation axis the dimensions get reduced to $8\times8\times64$. Since the invariant transition is integrated instead of average operation  \cite{b3}, a fully-connected layer adds $1k$ additional parameters to the existing network. The test error(\%) obtained by the existing and proposed models is shown in Table \ref{tab3}. Here, the proposed G-CNN+ZMs, G-CNN+PZMs and G-CNN+OFMMs reduces the test error significantly as compared to G-CNN. The lowest test error is achieved by G-CNN+PZMs of $6.90\%$, followed by G-CNN+ZMs of $6.98\%$ and G-CNN+OFMMs of $7.02\%$. Furthermore, CIFAR10+ dataset is generated by augmenting the horizontal flips and small translations to evaluate the impact of data augmentation\cite{b3}. The obtained results are shown in Table \ref{tab3}. Here also the proposed G-CNN+PZMs, G-CNN+ZMs and G-CNN+OFMMs performs significantly better than CNN and G-CNN.
\begin{table}[htbp]
\caption{Test error (\%) obtained on CIFAR10 and CIFAR10+ datasets.}
\begin{center}
\begin{tabular}{c|c|c|c}
\hline
\textbf{Method} & \textbf{\textit{CIFAR10}}& \textbf{\textit{CIFAR10+}}&\textbf{\textit{\#params}}\\
\hline
CNN \cite{b3} & 9.44 &8.86 & 1.37M\\ 
G-CNN \cite{b3} & 7.59 &7.04& 1.22M\\ 
\hline
G-CNN+ZMs & 6.98$\pm$0.007 &6.45$\pm$0.005&1.23M \\ 
G-CNN+PZMs & 6.90$\pm$0.004 &6.40$\pm$0.003&1.23M \\
G-CNN+OFMMs & 7.02$\pm$0.009&6.52$\pm$0.007 &1.23M \\ \hline
\end{tabular}

\label{tab3}
\end{center}
\end{table}
\section{Conclusion}
\label{sec8}
We have proposed the integration of ORIMs in the transition between convolution and fully-connected layers to learn the invariant representation in equivariant CNNs (e.g. G-CNNs). The mathematical framework of ORIMs for equivariant CNNs which is invariant to rotation, reflection and translation is also provided. The experiments are performed using  Zernike, pseudo-Zernike and orthogonal Fourier-Mellin moments. Our experimental results on rotated MNIST and CIFAR10 datasets show that the proposed integration of ORIMs improve the performance of equivariant CNNs (i.e., G-CNNs) significantly. Thus, the proposed invariant transition using ORIMs can be used in equivariant CNN architectures as a replacement to the down-sampling layers which are only used to remove the spatial dimensions of the filter responses to get the scalar fields for classification. Among  G-CNN+ZMs, G-CNN+PZMs and G-CNN+OFMMs, G-CNN+PZMs achieves the lowest test error(\%) followed by G-CNN+ZMs and G-CNN+OFMMs. The kernel functions of ZMs and  PZMs have similar characteristics while PZMs provide twice number of moments as compared to ZMs. Since, PZMs have more lower-order moments for same order $p_{max}$ than ZMs, thus, PZMs are less sensitive to transformations in an input function which is the reason PZMs performed well. OFMMs are useful for small functions because they provide more moment coefficients than ZMs and PZMs \cite{b16}. Here, OFMMs are conditioned to be $|q|\leq p$, provides same number of moment coefficients as PZMs, thus, provides less improvement.

\appendix
\subsection{Rotation Invariants}
\label{App:A}
Detailed proof of rotation invariance (Section \ref{RRI}). Let $O^\alpha (r,\theta)$ is the rotated version of function $O(r,\theta)$, rotated by angle $\alpha$ in the counter clockwise direction then the moment function $OM_{p,q}(O^\alpha)$ is defined as follows \cite{b5,b12,b17}:
\begin{equation}\tag{A1} \label{eq1}
\begin{split}
OM_{p,q} (O^\alpha ) & =\lambda_p\int_{0}^{2\pi}\int_{0}^{1}O^{\alpha}(r,\theta)R_{p,q}(r)e^{-iq\theta}rdrd\theta,\\
& = \lambda_p\int_{0}^{2\pi}\int_{0}^{1}O(r,\theta+\alpha)R_{p,q}(r)e^{-iq\theta}rdrd\theta, \\
& = \lambda_p\int_{0}^{2\pi}\int_{0}^{1}O(r,\theta ^{'})R_{p,q}(r)e^{-iq(\theta ^{'}-\alpha)}rdrd\theta ^{'},\\
& = \lambda_p\int_{0}^{2\pi}\int_{0}^{1}O(r,\theta ^{'})R_{p,q}(r)e^{-iq\theta ^{'}}e^{iq\alpha}rdrd\theta ^{'},\\
& = e^{iq\alpha}OM_{p,q}(O).\\
\end{split}
\end{equation}
This relationship shows that the moments of the original and the rotated images undergo phase-shift by an angle $q\alpha$ and the magnitude of the moments remain the same.


\subsection{Reflection Invariants}
\label{App:B}
Detailed proof of reflection invariance (Section \ref{RI}). Let $O^{hf} (s,t )=O(-s,t)$ is the horizontal flipped version of the discrete function $O(s,t)$ then the ORIMs are defined as follows \cite{b12}:
\begin{equation}\tag{A2} 
\begin{split}
OM_{p,q}(O^{hf}) & =\lambda_p\sum_{s=0}^{M-1}\sum_{t=0}^{M-1}O(-x_s,y_t)R_{p,q}(r_{s,t})e^{-iq\theta_{st}},\\
& =\lambda_p\sum_{s=0}^{M-1}\sum_{t=0}^{M-1}O(x_s,y_t)R_{p,q}(r_{s,t})e^{-iq(\pi-\theta_{st})},\\
& =\lambda_p\sum_{s=0}^{M-1}\sum_{t=0}^{M-1}(-1)^q O(x_s,y_t)R_{p,q}(r_{s,t})e^{iq\theta_{st}},\\
& = (-1)^q OM_{p,q}^{\ast} (O).\\
\end{split}
\end{equation}
Similarly, for vertical flipped $O^{vf} (s,t)=O(s,-t)$
\begin{equation}\tag{A3} 
\begin{split}
OM_{p,q}(O^{vf}) & =\lambda_p\sum_{s=0}^{M-1}\sum_{t=0}^{M-1}O(x_s,-y_t)R_{p,q}(r_{s,t})e^{-iq\theta_{st}},\\
& =\lambda_p\sum_{s=0}^{M-1}\sum_{t=0}^{M-1}O(x_s,y_t)R_{p,q}(r_{s,t})e^{-iq(-\theta_{st})},\\
& =\lambda_p\sum_{s=0}^{M-1}\sum_{t=0}^{M-1} O(x_s,y_t)R_{p,q}(r_{s,t})e^{iq\theta_{st}},\\
& = OM_{p,q}^{\ast} (O),
\end{split}
\end{equation}
where $OM_{p,q}^{\ast} (O)$ is the complex conjugate of $OM_{p,q} (O)$.

\subsection{Translation Invariants}
\label{App:C}
\renewcommand{\thefigure}{A1}
\begin{figure}
\begin{center}
\centerline{\includegraphics[width=8.0cm]{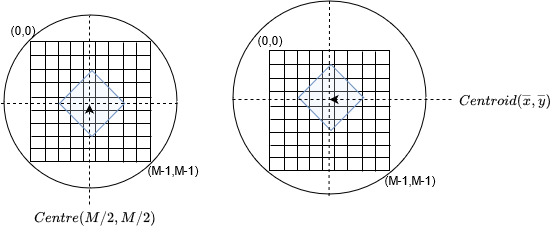}}
\caption{Mapping of unit disk to the center $(M/2,M/2)$ of an image (left) and mapping of unit disk to the centroid $(\overline{x},\overline{y})$ of translated image(right).}
\label{fig:A1}
\end{center}
\end{figure}
Detailed proof of translation invariance (Section \ref{TI}). Let $O'(s', t')$ is the translated version of discrete function $O(s,t)$, translated by $\Delta s$ and $\Delta t$ in the direction $s$ and $t$, respectively. The central moments $\overline{OM}_{p,q}(O')$ in (12) are computed using \eqref{seq:4} instead of (6) which replaces the digital center $(M/2,N/2)$ of $O'$ by its centroid $(\overline{x}, \overline{y})$ \cite{b17} as follows:
\begin{equation}\tag{A4}\label{seq:4}
    \overline{x}_s = \frac{2s+1-\overline{x}}{D},\
    \overline{y}_t = \frac{2t+1-\overline{y}}{D}.
\end{equation}
The centroid $(\overline{x}, \overline{y})$ are obtained as follows \cite{b12}:
\begin{equation}\tag{A5}\label{seq:5}
    \overline{x} = \frac{\sum_{s=0}^{M-1}\sum_{t=0}^{M-1}s.O(s,t)}{\sum_{s=0}^{M-1}\sum_{t=0}^{M-1}O(s,t)},\
    \overline{y} = \frac{\sum_{s=0}^{M-1}\sum_{t=0}^{M-1}t.O(s,t)}{\sum_{s=0}^{M-1}\sum_{t=0}^{M-1}O(s,t)}.
\end{equation}
Figure \ref{fig:A1} shows the mapping of unit disk to the center $(M/2,M/2)$ and centroid $(\overline{x},\overline{y})$ of an image where \eqref{seq:5} is used to compute the centroids.
\end{document}